%% file: Paper.tex
\documentclass[10pt,twocolumn,letterpaper]{article}

\usepackage{cvpr}              

\input{preamble}

\definecolor{cvprblue}{rgb}{0.21,0.49,0.74}
\usepackage[pagebackref,breaklinks,colorlinks,allcolors=cvprblue]{hyperref}

\def\paperID{*****} 
\def\confName{CVPR}
\def\confYear{2026}

\title{Data-Efficient Surgical Phase Segmentation in Small-Incision Cataract Surgery: A Controlled Study of Vision Foundation Models}

\author{Lincoln Spencer, Song Wang, Chen Chen\\
Institute of Artificial Intelligence, University of Central Florida\\
}

\begin{document}
\maketitle
\input{0_abstract}
\input{1_intro}
\input{2_method}
\input{3_experiments}
\input{4_results}
\FloatBarrier
\input{5_conclusion}
\FloatBarrier
{
    \small
    \bibliographystyle{ieeetr}
    \bibliography{main}
}


\end{document}

%% file: preamble.tex
\usepackage{graphicx}
\usepackage{booktabs}
\usepackage{multirow}
\usepackage{amsmath,amssymb}
\usepackage{microtype}
\usepackage{xspace}
\usepackage{colortbl}
\usepackage{siunitx}
\usepackage{enumitem}
\usepackage{url}
\usepackage{placeins}

\graphicspath{{./}}

\newcommand{\dataset}{SICS-155\xspace}
\newcommand{\mstcn}{MS-TCN++\xspace}
\newcommand{\dinovthree}{DINOv3\xspace}
\newcommand{\vjepa}{V-JEPA2\xspace}

\makeatletter
\@ifundefined{linenumbersep}{}{\setlength{\linenumbersep}{0.30cm}}
\makeatother

\definecolor{bestcell}{RGB}{245,245,245}

%% file: 0_abstract.tex
\begin{abstract}
Surgical phase segmentation is central to computer-assisted surgery, yet robust models remain difficult to develop when labeled surgical videos are scarce. We study data-efficient phase segmentation for manual small-incision cataract surgery (SICS) through a controlled comparison of visual representations. To isolate representation quality, we pair each visual encoder with the same temporal model (MS-TCN++) under identical training and evaluation settings on SICS-155 (19 phases). We compare supervised encoders (ResNet-50, I3D) against large self-supervised foundation models (DINOv3, V-JEPA2), and use a cached-feature pipeline that decouples expensive visual encoding from lightweight temporal learning. Foundation-model features improve segmentation performance in this setup, with DINOv3 ViT-7B achieving the best overall results (83.4\% accuracy, 87.0 edit score). We further examine cataract-domain transfer using unlabeled videos and lightweight adaptation, and analyze when it helps or hurts. Overall, the study indicates strong transferability of modern vision foundation models to surgical workflow understanding and provides practical guidance for low-label medical video settings. The project website is available at: \href{https://sl2005.github.io/DataEfficient-sics-phase-seg/}{https://sl2005.github.io/DataEfficient-sics-phase-seg/}
\end{abstract}

%% file: 1_intro.tex
\section{Introduction}
Surgical workflow understanding from video can support intraoperative assistance, automatic documentation, skill assessment, and safety monitoring. In practice, however, progress is often limited by annotation cost: phase segmentation requires dense frame-level labels over long procedures, which are expensive to obtain at scale~\cite{ward2021annotation}. This challenge is especially pronounced for \emph{manual small-incision cataract surgery} (SICS), which is widely performed in low-resource settings where annotation capacity and specialized infrastructure may be limited~\cite{tabin2008developingworld,venkatesh2012msics}.

Phase segmentation is also intrinsically a joint spatial--temporal task. Strong performance depends not only on recognizing instruments and anatomy in individual frames, but also on modeling long-range workflow structure, repeated sub-steps, and brief transitions between clinically similar phases.

Motivated by these constraints, we position this work as a \emph{systematic controlled study} of data-efficient surgical video understanding, rather than as a new temporal architecture. We leverage self-supervised visual foundation models pretrained on large unlabeled corpora to provide strong frame-level features with limited task-specific supervision. To keep comparisons interpretable and deployment-relevant, we fix the temporal learner to \mstcn~\cite{ms_tcnpp} and vary only the feature extractor. This design (i) isolates representation quality by controlling temporal architecture, optimization, and splits, and (ii) matches a practical workflow where heavy encoder inference is performed once offline (feature caching) and the temporal head is trained quickly on limited labels. Section~\ref{sec:experiments} details this shared protocol.

The contributions of this paper are summarized as follows.
\begin{itemize}[leftmargin=1.25em,topsep=0.2em,itemsep=0.2em]
\item We present a controlled benchmark for surgical phase segmentation where supervised encoders and self-supervised foundation models are evaluated under the same cached-feature workflow, shared \mstcn head, and stratified splits.
\item We study empirical representation scaling and deployment trade-offs across encoder families, while carefully interpreting scaling trends because model size and upstream pretraining scale co-vary in released checkpoints.
\item We analyze \emph{CataractFT}, an exploratory cataract-domain transfer workflow combining self-supervised continuation and optional LoRA adaptation before frozen feature extraction on \dataset~\cite{sics155}, and characterize when transfer is helpful versus harmful.
\end{itemize}

\vspace{0.5em}
\noindent\textbf{Practical insight.} For teams building surgical video systems with limited labels, foundation-model features can be a strong drop-in alternative to supervised encoders when the temporal model is held fixed and evaluation includes both frame-level and segment-level metrics.

\paragraph{Small Data Statement.}
This study is \emph{small data} in two senses. First, \dataset contains only 155 videos, and each sample requires dense frame-level labeling across 19 phases; for rare phases, the effective sample size is much smaller than the raw video count. Second, the target domain is operationally narrow (manual SICS in low-resource settings), making large, task-matched labeled corpora difficult and costly to assemble.

To address this setting, we explicitly design for data efficiency. We reuse representations from large self-supervised visual encoders pretrained on external unlabeled data, rather than training video models from scratch on SICS. We keep the temporal backbone fixed so improvements can be attributed to representation quality, use stratified 5-fold cross-validation to reduce split sensitivity, cache features to enable efficient temporal training, and explore CataractFT (self-supervised continuation with optional parameter-efficient adaptation on unlabeled cataract-domain video) as a practical way to exploit additional data without requiring new SICS labels.

%% file: 2_method.tex
\section{Method}
\subsection{Problem setup}
Given a surgical video with $T$ frames, the goal is to assign a phase label $y_t \in \{1,\dots,C\}$ to each frame $t$. We study how the per-frame representation $\mathbf{f}_t \in \mathbb{R}^d$ affects downstream temporal segmentation.

\subsection{Overall Framework}
End-to-end training of very large visual encoders on long surgical videos is often computationally prohibitive and prone to overfitting, especially on long-tailed datasets like \dataset. Instead of proposing a new temporal architecture, we adopt a deployment-oriented factorization: a large visual encoder produces frozen per-frame features, and a lightweight temporal model maps cached features to phase labels. This decoupling enables controlled comparisons, reduces downstream training cost, and matches realistic deployment where expensive feature extraction is run once offline.

Our \textbf{base benchmark pipeline} caches features from off-the-shelf encoders and trains a shared \mstcn head on those tensors. We then study \emph{CataractFT} as an \textbf{exploratory extension}: optional self-supervised continuation on 1{,}000 unlabeled Cataract-1K videos using each backbone's native objective (DINO self-distillation or JEPA masked prediction), followed by optional LoRA phase-classification adaptation~\cite{lora} on 56 annotated Cataract-1K and 101 Cataract-101 surgeries~\cite{cataract1k,cataract101}. In our implementation, LoRA uses rank $r=16$ and $\alpha=32$ for up to 50 epochs. The adapted encoder is then frozen for dense SICS feature extraction, and only the same \mstcn head is trained on \dataset.

\begin{figure*}[!t]
    \centering
    \includegraphics[width=0.96\textwidth]{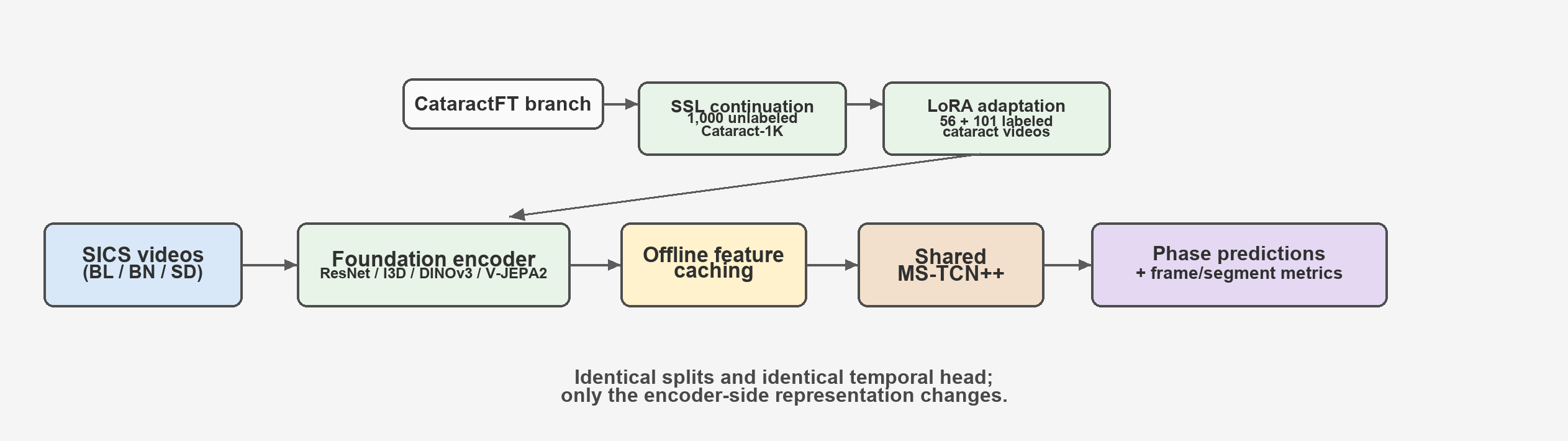}
    \caption{Controlled study pipeline. Encoders are frozen after optional cataract-domain SSL continuation and LoRA adaptation, features are cached once, and the same \mstcn head is trained for all encoders.}
    \label{fig:pipeline-overview}
\end{figure*}

Figure~\ref{fig:pipeline-overview} summarizes the controlled pipeline, including the shared cached-feature benchmark path and the optional CataractFT transfer branch.

\subsection{Feature extractors}
We compare two encoder groups:
\begin{itemize}[leftmargin=*]
    \item \textbf{Supervised encoders:} ResNet-50 pretrained on ImageNet~\cite{imagenet} and I3D~\cite{i3d} pretrained on Kinetics.
    \item \textbf{Self-supervised foundation models:} \dinovthree~\cite{dinov3} (image ViTs) and \vjepa~\cite{vjepa2} (video ViTs).
\end{itemize}

\paragraph{Frame-level features (\dinovthree).}
For \dinovthree, we extract one feature per frame with a single forward pass of the image encoder and use the pooled representation as $\mathbf{f}_t$.

\paragraph{Clip-to-frame features (\vjepa).}
\vjepa operates on clips (e.g., 64 frames). To obtain per-frame features, we use a sliding-window procedure and assign each clip embedding to its center frame. For tractability (especially ViT-g at 384px), we extract at stride 4 and linearly interpolate back to full frame rate before training \mstcn; this may smooth rapid transitions.

Implementation details for feature caching, optimization, and the fixed \mstcn configuration are deferred to Section~\ref{sec:implementation}.

%% file: 3_experiments.tex
\section{Experiments}
\label{sec:experiments}
\subsection{Dataset}
\begin{figure*}[!tb]
    \centering
    \includegraphics[width=0.98\textwidth,height=0.4\textheight,keepaspectratio]{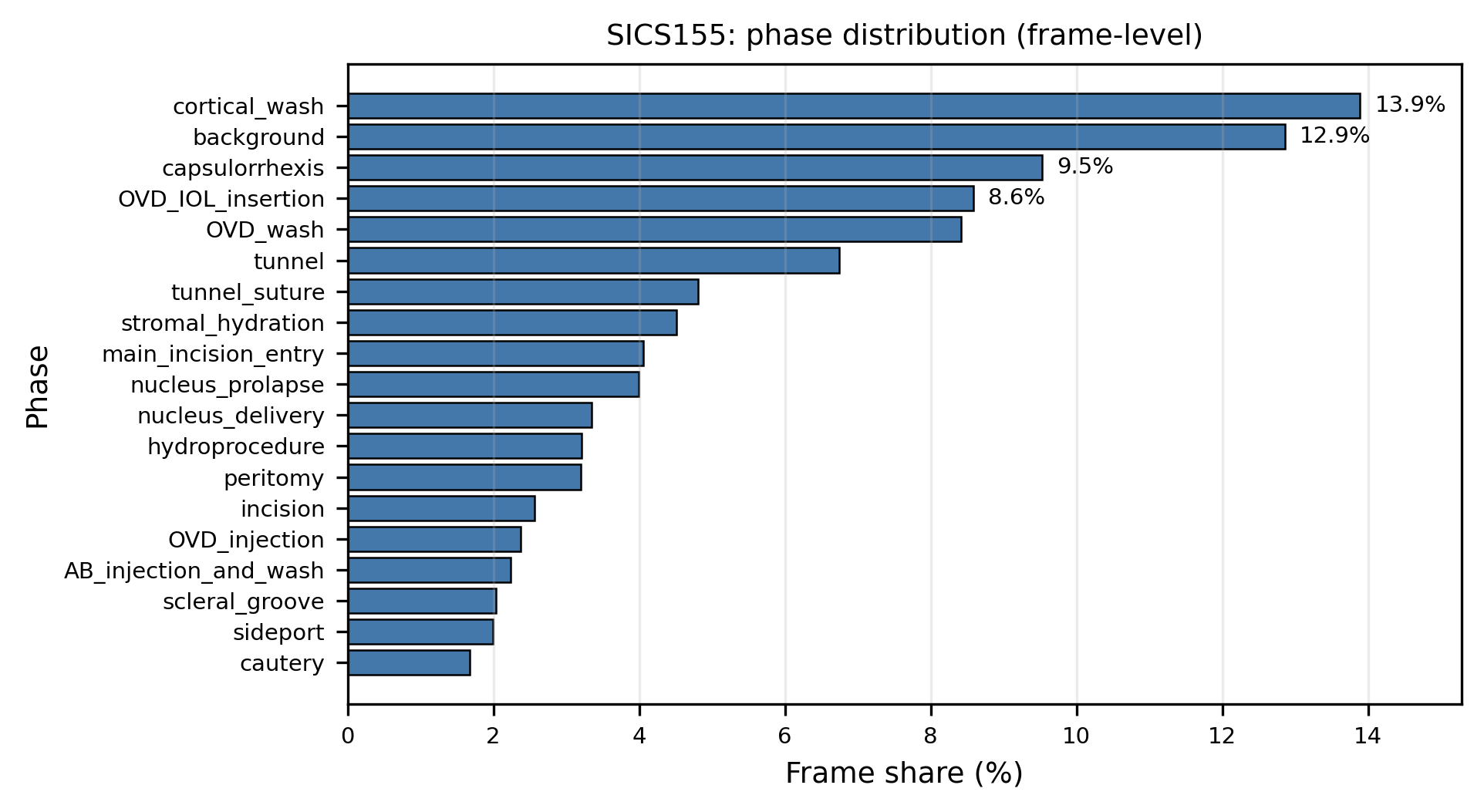}
    \caption{\dataset phase distribution at frame level. The long-tail imbalance motivates representation learning approaches that remain robust under limited labels.}
    \label{fig:phase-dist}
\end{figure*}

We evaluate on \dataset~\cite{sics155}, which contains 155 annotated SICS procedures with frame-level labels over 19 phases. The released train/validation portion available for this study contains 115 labeled videos (32 BL, 50 BN, 33 SD), where BL, BN, and SD denote the three procedure-type prefixes used in the released annotations; the hidden test set is not used. This public subset already reflects key dataset challenges: median video duration is 6.4 minutes (range 2.5--15.0), \emph{cautery} accounts for only 1.7\% of frames and appears in 48/115 videos, \emph{cortical wash} occupies 13.8\% of frames and appears in every video, and short phases such as \emph{sideport} and \emph{scleral groove} have median contiguous durations of about 6.5 seconds. Because only one released annotation stream is available per video, inter-rater annotation noise cannot be quantified directly and remains a limitation. Figure~\ref{fig:phase-dist} shows the frame-level class imbalance.

Table~\ref{tab:encoder-specs} summarizes the evaluated backbones, modalities, input resolutions, and feature dimensionalities used throughout the benchmark.

\begin{table*}[!tb]
\centering
\caption{Feature extractor backbones used in our pipeline. ``Feat dim'' is the per-frame feature dimensionality consumed by \mstcn. Parameter counts are for the encoder backbone only; V-JEPA2/ResNet/I3D/DINOv3-B are computed from instantiated architectures (without loading weights), while DINOv3-L and DINOv3-7B are from model releases. V-JEPA2 uses stride-4 clip extraction with linear interpolation to full frame rate.}
\label{tab:encoder-specs}
\small
\renewcommand{\arraystretch}{0.99}
\setlength{\tabcolsep}{3pt}
\begin{tabular}{l l l c c c r}
\toprule
Encoder & Pretraining / Source & Modality & Input & Clip & Feat dim & Params \\
\midrule
ResNet-50 (ImageNet) & supervised (ImageNet-1K) & 2D image & $224$ & -- & $2048$ & $25.6$M \\
I3D-R50 (Kinetics) & supervised (Kinetics-400) & 3D video & $224$ & $16$ & $2048$ & $28.0$M \\
DINOv3 ViT-B/16 & self-supervised (LVD-1689M) & 2D image & $224$ & -- & $768$ & $85.7$M \\
DINOv3 ViT-L/16 & self-supervised (LVD-1689M) & 2D image & $224$ & -- & $1024$ & $300$M \\
DINOv3 ViT-7B/16 & self-supervised (LVD-1689M) & 2D image & $224$ & -- & $4096$ & $6.716$B \\
V-JEPA2 ViT-L & self-supervised video (Meta) & 3D video & $256$ & $64$ & $1024$ & $326.0$M \\
V-JEPA2 ViT-g/16 & self-supervised video (Meta) & 3D video & $384$ & $64$ & $1408$ & $1.03$B \\
\bottomrule
\end{tabular}
\end{table*}

\subsection{Evaluation protocol and controlled design}
We use stratified 5-fold cross-validation with folds balanced by procedure type. Folds are constructed from the union of the available training and validation videos, and we report mean$\pm$std over five runs.

Our primary question is representation quality: with the same temporal model, does changing only the feature extractor improve phase segmentation? To answer this, we enforce a controlled protocol. Across all experiments, the temporal learner is fixed to \mstcn~\cite{ms_tcnpp}, which takes $\mathbf{F}=[\mathbf{f}_1,\dots,\mathbf{f}_T]$ and predicts per-frame phase probabilities. We freeze each encoder and keep architecture settings, optimization, number of epochs, loss terms, and splits identical across runs. This minimizes confounds and makes performance differences directly attributable to feature quality rather than temporal-head retuning. The trade-off is that we do \emph{not} measure how stronger temporal architectures might change absolute scores or relative gaps.

\subsection{Metrics}
To cover both deployment needs and temporal segmentation practice, we report:
\begin{itemize}[leftmargin=*]
    \item \textbf{Frame-level:} accuracy, macro-F1, PR-AUC.
    \item \textbf{Segment-level:} edit score (normalized Levenshtein, rewarding correct ordering while penalizing spurious segments), and segmental $\mathrm{F1}@\{10,25,50\}$ at IoU thresholds.
    \item \textbf{Overlap:} mIoU (Jaccard index).
\end{itemize}

\subsection{Implementation details}
\label{sec:implementation}
All downstream experiments share the same splits, feature-caching workflow, temporal model, and optimization; only the frozen feature extractor changes. We generate stratified 5-fold splits once from the official train+validation pool, balancing by procedure prefix (BL/BN/SD), and report mean$\pm$std over held-out folds. All runs use the same random seed and deterministic cuDNN settings.

For each encoder, we first cache per-video feature tensors offline, then train only the temporal head on those embeddings. Feature extraction is aligned to each video's ground-truth frame count. ResNet-50 uses ImageNet preprocessing~\cite{imagenet} (resize 256, center crop 224); I3D uses 16-frame clips at $224\times224$ with Kinetics normalization; DINOv3 uses pooled per-frame image features; and V-JEPA2 uses 64-frame clips at $256^2$ (ViT-L) or $384^2$ (ViT-g). For tractability on long videos, V-JEPA2 features are extracted every fourth frame and linearly interpolated to full annotation rate before temporal training. CataractFT variants load the best saved LoRA adapter~\cite{lora} for the corresponding backbone during extraction, after which the encoder remains frozen. If minor feature-label length mismatches remain, both streams are truncated to their minimum shared length.

The temporal model is identical for all encoders, following the fixed \mstcn~configuration used in prior SICS phase-recognition work~\cite{ms_tcnpp,mueller2025sics105}: 13 prediction-generation layers, 4 refinement stages, 13 layers per refinement stage, and 64 hidden feature maps. We train for 100 epochs with batch size 1 using Adam at learning rate $5\times10^{-4}$, with a MultiStep schedule multiplying the learning rate by $0.3$ at 60\% and 90\% of training. The loss is stage-wise cross-entropy plus a temporal smoothing MSE term (weight 0.35); adaptive smoothing uses a 30-frame window, while Dice and focal losses are disabled in reported experiments. Each fold is evaluated using the final-epoch checkpoint, and we save frame labels and per-frame class probabilities for PR-AUC computation. All experiments were run on a single NVIDIA B200 GPU, reflecting our emphasis on small-data, deployment-relevant adaptation.

%% file: 4_results.tex
\section{Results}

\subsection{Cross-model comparison}
Table~\ref{tab:kfold-results} reports 5-fold cross-validation results across encoders.
\input{kfold_table_paper.tex}
Self-supervised \dinovthree features outperform ResNet-50 on average and are broadly comparable to I3D in this controlled setup; for close pairs (e.g., I3D vs. DINOv3 ViT-B), differences should be interpreted cautiously given fold-to-fold variability (paired fold-level tests are available on our project website). For the primary supervised-versus-foundation comparison, DINOv3 ViT-L and ViT-7B show higher accuracy, macro-F1, and edit score than I3D in all five folds. We also observe a consistent within-family trend for \dinovthree (ViT-B $<$ ViT-L $<$ ViT-7B). Because released checkpoints differ in both parameter count and upstream pretraining scale, we interpret this as an empirical scaling observation within one model family, not a causal claim about parameter count alone.

\begin{figure*}[!tb]
	\centering
	\includegraphics[width=0.99\textwidth,height=0.7\textheight,keepaspectratio]{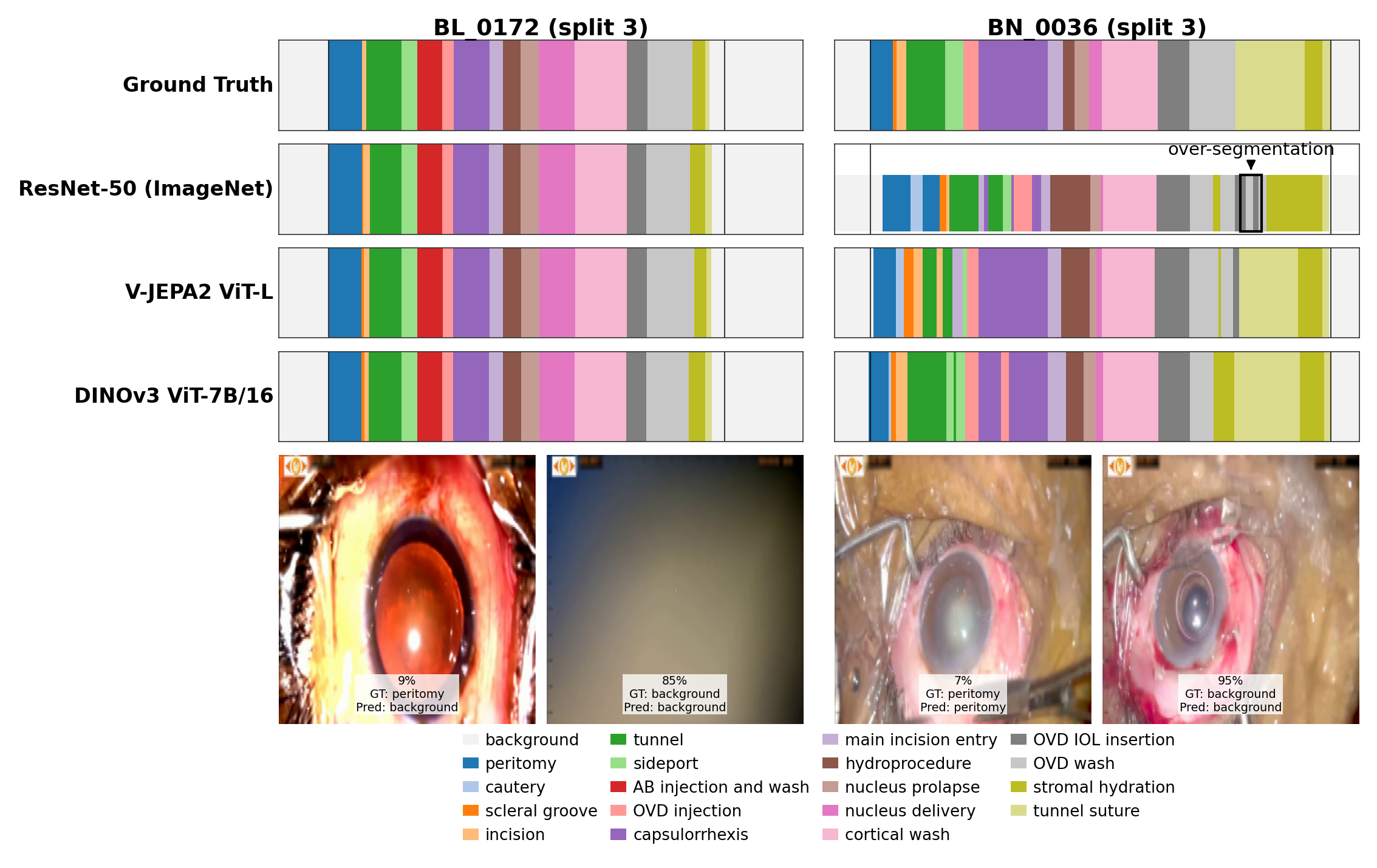}
	\caption{Qualitative phase segmentation results for BL\_0172 and BN\_0036 (colors defined by the legend). For each video, rows show ground truth, the supervised baseline (ResNet-50 features), a representative foundation-model alternative (V-JEPA2 ViT-L features), and the best-performing foundation model (DINOv3 ViT-7B features), all with the same temporal head. The bottom strip in each column shows aligned keyframes with sampled timepoints and corresponding ground-truth/predicted phases. The callout highlights a choppy baseline region that is reduced with stronger representations.}
	\label{fig:qualitative-ribbons}
\end{figure*}

Figure~\ref{fig:qualitative-ribbons} provides representative qualitative examples for two procedures and illustrates how stronger representations reduce visually choppy predictions relative to the supervised baseline.

We include both ResNet-50 and I3D as supervised baselines: I3D is commonly used in prior surgical workflow work, while ResNet-50 provides a simple 2D reference when temporal reasoning is delegated entirely to the shared MS-TCN++ head. Results should be interpreted in light of the controlled design in Section~\ref{sec:experiments}: fixing \mstcn isolates encoder quality, but stronger temporal heads could change absolute performance and possibly narrow or widen some gaps at difficult boundaries.

\paragraph{Why does V-JEPA2 underperform despite being video-native?}
V-JEPA2 uses 64-frame clips; for tractable extraction on long videos (especially ViT-g at 384px), we compute clip embeddings at stride 4 and linearly interpolate to full frame rate. This may smooth rapid transitions and weaken boundary cues, which can reduce segment-level metrics (edit and $\mathrm{F1}@k$). Domain shift from generic pretraining video is a second plausible factor.

\subsection{Domain adaptation via cataract transfer (CataractFT)}
In emerging deployments, teams may have access to \emph{unlabeled} videos from related cataract procedures (e.g., phacoemulsification) but limited labeled SICS data. Our repository includes a practical pipeline for this setting: self-supervised continuation on 1{,}000 unlabeled Cataract-1K videos, optional LoRA fine-tuning on 56 annotated Cataract-1K plus 101 Cataract-101 surgeries, and frozen feature extraction for the same SICS temporal head.

In this preliminary study, transfer is mixed and can be strongly negative (Table~\ref{tab:cataractft-transfer}), despite substantial phase overlap with SICS (\mbox{$\approx$68.6\%} of non-background frames map to phases with direct phaco analogues under a conservative mapping).

\input{cataractft_transfer_table_paper.tex}
\input{cataractft_ablation_table_paper.tex}

\begin{figure*}[!t]
	\centering
	\includegraphics[width=0.85\textwidth]{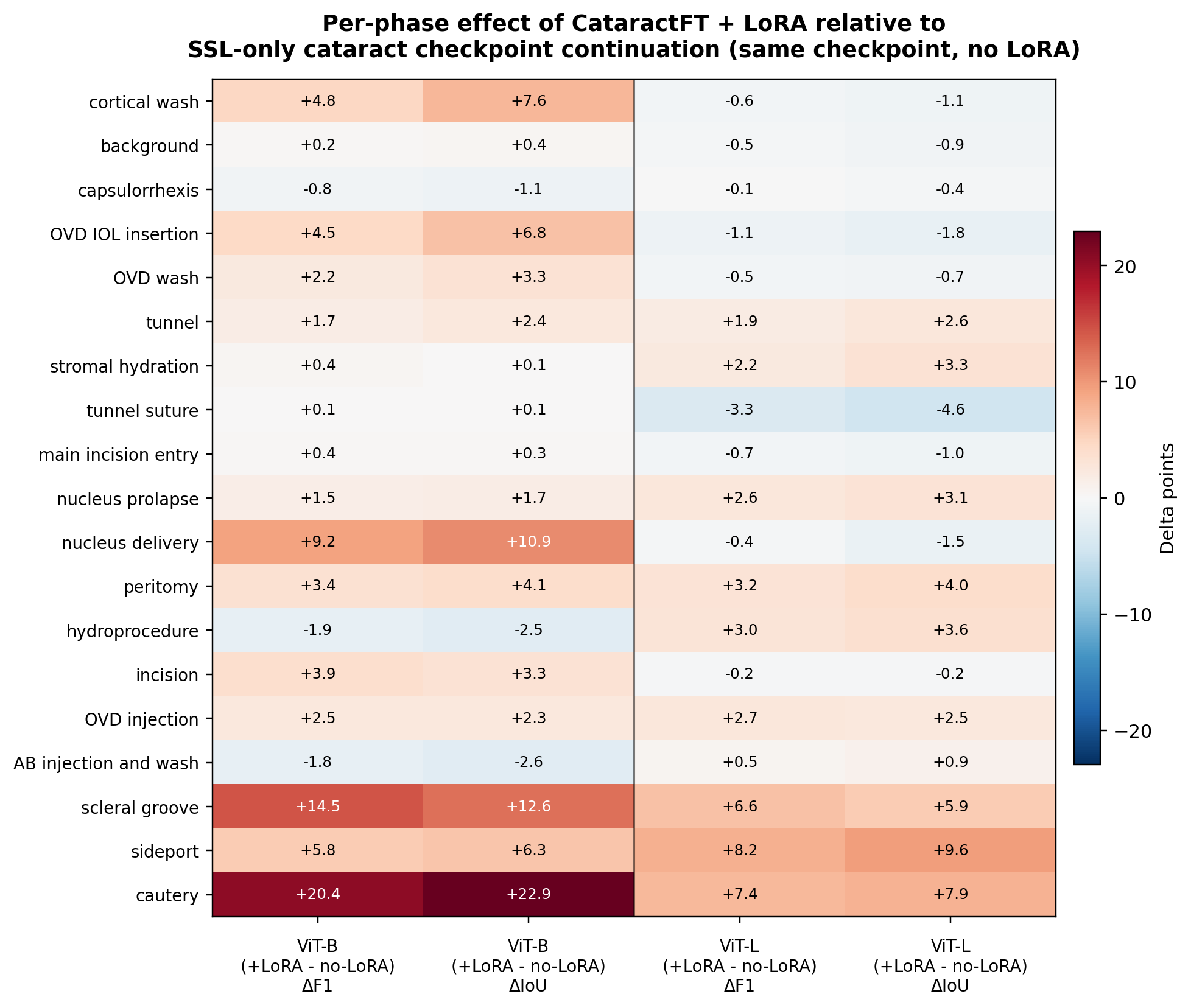}
	\caption{Per-phase change from CataractFT + LoRA relative to SSL-only continuation from the same cataract-domain checkpoint (no LoRA). Heatmap entries report delta points for classwise F1 and IoU for DINOv3 ViT-B and ViT-L. Positive values indicate phases that benefit from LoRA adaptation; negative values indicate phases where the checkpoint-only variant remains better.}
	\label{fig:cataractft-phase-delta}
\end{figure*}

\paragraph{SSL-only checkpoint continuation vs. CataractFT + LoRA.}
Comparing full CataractFT against an SSL-only variant initialized from the same cataract-domain checkpoint but \emph{without} LoRA reveals a consistent trade-off. For both DINOv3 backbones, CataractFT + LoRA improves frame-centric performance on \dataset. For ViT-B, accuracy, macro-F1, PR-AUC, and mIoU increase by $+3.01$, $+3.65$, $+2.29$, and $+4.15$ points; for ViT-L, gains are $+0.46$, $+1.42$, $+0.85$, and $+1.65$. Segmental overlap also improves slightly at stricter thresholds (e.g., $\mathrm{F1}@50$ rises by $+1.75$ for ViT-B and $+1.10$ for ViT-L).

However, edit score decreases for both backbones relative to SSL-only continuation ($-1.29$ for ViT-B, $-1.80$ for ViT-L). This indicates sharper framewise discrimination without consistently smoother temporal sequences: predictions can become more accurate overall while introducing short extra segments or boundary oscillations that edit penalizes strongly. A second pattern is that ViT-B benefits more than ViT-L from CataractFT, consistent with smaller backbones having more room for task-adjacent specialization.

Table~\ref{tab:cataractft-ablation} adds an important nuance. For ViT-B, full CataractFT + LoRA is strongest on most frame-level metrics and remains above the original base extractor on edit score. For ViT-L, both cataract-domain variants trail the original base extractor on aggregate accuracy, macro-F1, PR-AUC, F1@50, and mIoU, while SSL-only continuation is best only for edit score. This suggests backbone-dependent adaptation behavior: checkpoint continuation may help temporal consistency, whereas LoRA is most beneficial when the starting representation has more headroom to specialize.

Figure~\ref{fig:cataractft-phase-delta} shows that these effects are phase-dependent rather than uniform. Gains concentrate in clinically important phases such as tunnel suture, sideport, and cautery, while some classes (e.g., cortical wash for ViT-L and hydroprocedure for ViT-B) change little or regress. We hypothesize this reflects differences in visual signatures. LoRA appears to help detection of specific localized instruments (e.g., cautery probes), but is less effective for amorphous phases such as cortical wash that require stronger long-range temporal context. Because our pipeline keeps the temporal head frozen, improved spatial discrimination can over-emphasize local frame cues without equivalent temporal recalibration, producing fragmented predictions near phase boundaries.

\noindent\textbf{Deployment note.}
ViT-L may offer a better accuracy/compute trade-off for deployment; for context, ViT-7B weights alone require \mbox{$\approx$13.4~GB} of VRAM in bf16/fp16 (excluding activations/overheads).

%% file: kfold_table_paper.tex
\begin{table*}[!tb]
\centering
\caption{5-Fold Cross Validation Results on \dataset (mean $\pm$ std, \%). Best per metric is bolded and lightly shaded.}
\label{tab:kfold-results}
\small
\renewcommand{\arraystretch}{0.99}
\setlength{\tabcolsep}{4pt}
\begin{tabular}{lcccccccc}
\toprule
Model & Accuracy & F1 Score (macro) & Edit Score & PR-AUC & F1@10 & F1@25 & F1@50 & mIoU (Jaccard) \\
\midrule
ResNet-50 (ImageNet) & 75.7 $\pm$ 1.1 & 67.7 $\pm$ 1.4 & 79.3 $\pm$ 1.5 & 71.8 $\pm$ 1.2 & 79.8 $\pm$ 1.1 & 75.3 $\pm$ 2.0 & 62.2 $\pm$ 2.5 & 53.5 $\pm$ 1.6 \\
I3D & 79.8 $\pm$ 0.9 & 71.7 $\pm$ 1.4 & 82.6 $\pm$ 2.5 & 75.7 $\pm$ 1.1 & 83.2 $\pm$ 2.5 & 79.2 $\pm$ 3.2 & 68.9 $\pm$ 2.6 & 58.3 $\pm$ 1.8 \\
DINOv3 ViT-B/16 & 78.4 $\pm$ 0.7 & 71.2 $\pm$ 0.9 & 82.5 $\pm$ 1.7 & 75.0 $\pm$ 1.2 & 83.8 $\pm$ 1.1 & 79.9 $\pm$ 1.8 & 68.2 $\pm$ 2.0 & 57.0 $\pm$ 2.0 \\
DINOv3 ViT-L/16 & 82.2 $\pm$ 1.3 & 75.8 $\pm$ 2.1 & 85.7 $\pm$ 1.7 & 78.5 $\pm$ 1.3 & 87.2 $\pm$ 1.8 & 84.4 $\pm$ 2.5 & 73.7 $\pm$ 4.4 & 62.4 $\pm$ 3.6 \\
DINOv3 ViT-7B/16 & \cellcolor{bestcell}\textbf{83.4 $\pm$ 1.4} & \cellcolor{bestcell}\textbf{76.5 $\pm$ 2.0} & \cellcolor{bestcell}\textbf{87.0 $\pm$ 1.5} & \cellcolor{bestcell}\textbf{79.4 $\pm$ 1.7} & \cellcolor{bestcell}\textbf{88.0 $\pm$ 2.2} & \cellcolor{bestcell}\textbf{84.5 $\pm$ 2.6} & \cellcolor{bestcell}\textbf{75.0 $\pm$ 1.8} & \cellcolor{bestcell}\textbf{64.1 $\pm$ 3.2} \\
V-JEPA2 ViT-L & 77.9 $\pm$ 1.0 & 69.9 $\pm$ 0.5 & 83.2 $\pm$ 1.0 & 74.2 $\pm$ 0.6 & 83.0 $\pm$ 0.9 & 78.2 $\pm$ 1.5 & 66.6 $\pm$ 1.7 & 55.6 $\pm$ 2.3 \\
V-JEPA2 ViT-g & 76.0 $\pm$ 1.0 & 67.5 $\pm$ 1.2 & 81.2 $\pm$ 2.6 & 72.2 $\pm$ 0.7 & 80.6 $\pm$ 2.5 & 76.5 $\pm$ 2.2 & 63.8 $\pm$ 2.6 & 53.2 $\pm$ 2.1 \\
\bottomrule
\end{tabular}
\end{table*}

%% file: cataractft_transfer_table_paper.tex

\begin{table}[t]
\centering
\caption{Preliminary CataractFT transfer to \dataset. Deltas are measured on \dataset (5-fold CV) relative to the corresponding base encoder; \textbf{Acc (FT)} is the post-fine-tuning mean accuracy on \dataset.}
\label{tab:cataractft-transfer}
\small
\renewcommand{\arraystretch}{0.90}
\setlength{\tabcolsep}{2pt}
\begin{tabular}{p{0.40\columnwidth} r r r r}
\toprule
Encoder & Acc (FT) & $\Delta$Acc & $\Delta$Macro-F1 & $\Delta$Edit \\
\midrule
DINOv3 ViT-B/16 & 81.27 & +2.86 & +2.87 & +1.66 \\
DINOv3 ViT-L/16 & 81.26 & -0.93 & -1.82 & -0.69 \\
V-JEPA2 ViT-L & 71.53 & -6.37 & -8.08 & -3.31 \\
\bottomrule
\end{tabular}
\end{table}

%% file: cataractft_ablation_table_paper.tex
\begin{table*}[!tb]
\centering
\caption{Side-by-side CataractFT ablation for encoders with all three variants available on \dataset (5-fold CV, mean $\pm$ std, \%). \textbf{Base} uses the encoder as a standard frozen feature extractor, \textbf{SSL-only} continues from the same cataract-domain checkpoint without LoRA, and \textbf{CataractFT + LoRA} applies the full adaptation pipeline. Best within each encoder block is bolded.}
\label{tab:cataractft-ablation}
\scriptsize
\renewcommand{\arraystretch}{0.92}
\setlength{\tabcolsep}{4pt}
\begin{tabular}{l l c c c c c c}
\toprule
Encoder & Variant & Accuracy & Macro-F1 & Edit & PR-AUC & F1@50 & mIoU \\
\midrule
\multirow{3}{*}{DINOv3 ViT-B/16}
& Base & 78.41 $\pm$ 0.71 & 71.23 $\pm$ 0.87 & 82.55 $\pm$ 1.71 & 75.04 $\pm$ 1.23 & 68.20 $\pm$ 1.96 & 57.01 $\pm$ 1.96 \\
& SSL-only (no LoRA) & 78.26 $\pm$ 1.86 & 70.45 $\pm$ 2.19 & \textbf{85.50 $\pm$ 1.61} & 74.74 $\pm$ 1.66 & 68.83 $\pm$ 3.47 & 57.31 $\pm$ 2.20 \\
& CataractFT + LoRA & \textbf{81.27 $\pm$ 1.84} & \textbf{74.10 $\pm$ 2.23} & 84.21 $\pm$ 2.58 & \textbf{77.03 $\pm$ 2.02} & \textbf{70.58 $\pm$ 3.02} & \textbf{61.46 $\pm$ 2.61} \\
\midrule
\multirow{3}{*}{DINOv3 ViT-L/16}
& Base & \textbf{82.19 $\pm$ 1.28} & \textbf{75.80 $\pm$ 2.07} & 85.71 $\pm$ 1.74 & \textbf{78.50 $\pm$ 1.31} & \textbf{73.74 $\pm$ 4.44} & \textbf{62.40 $\pm$ 3.61} \\
& SSL-only (no LoRA) & 80.80 $\pm$ 1.31 & 72.56 $\pm$ 1.33 & \textbf{86.82 $\pm$ 0.56} & 77.08 $\pm$ 0.84 & 70.74 $\pm$ 1.51 & 59.58 $\pm$ 3.20 \\
& CataractFT + LoRA & 81.26 $\pm$ 0.91 & 73.98 $\pm$ 1.02 & 85.02 $\pm$ 2.04 & 77.93 $\pm$ 0.96 & 71.84 $\pm$ 3.22 & 61.23 $\pm$ 2.55 \\
\bottomrule
\end{tabular}
\end{table*}

%% file: 5_conclusion.tex
\section{Conclusion}
We present a controlled study of data-efficient surgical phase segmentation in MSICS under low-resource constraints. Using a shared \mstcn temporal head and frozen cached features, we show that self-supervised foundation-model representations provide clear overall gains over traditional supervised encoders in this benchmark, with \dinovthree~ViT-7B achieving the best overall performance.

At the same time, deployment decisions should not rely on peak accuracy alone. ViT-L may provide a more favorable accuracy/compute trade-off in realistic settings where MSICS is practiced with limited infrastructure. The observed scaling pattern should also be interpreted carefully, because model size and upstream pretraining scale are coupled in currently available checkpoints. CataractFT further suggests that related cataract-domain video can transfer useful clinical semantics to \dataset, but gains are mixed and backbone-dependent. In particular, adaptation can improve frame-level discrimination without consistently improving temporal consistency, reinforcing the need to report both frame-level and segment-level metrics.

Overall, the primary contribution is a systematic, controlled evaluation of foundation-model representations for surgical video understanding in low-label MSICS. CataractFT serves as a practical transfer-learning extension rather than the sole contribution. Future work should evaluate CataractFT more broadly across datasets and settings, better disentangle dependence on visible instruments versus broader operative context, and develop stronger temporal models and temporal-smoothness controls under adaptation.